\documentclass[11pt]{article}
\usepackage{acl2016}
\usepackage{times}
\usepackage{latexsym}
\usepackage{graphicx}
\usepackage{url}
\usepackage{amsmath}
\usepackage{amsfonts}
\usepackage{bm}

\aclfinalcopy 

\title{Composing Distributed Representations of Relational Patterns}

\author{Sho Takase \hspace{1em} Naoaki Okazaki \hspace{1em} Kentaro Inui \\
  Graduate School of Information Sciences, Tohoku University \\
  {\tt \{takase, okazaki, inui\}@ecei.tohoku.ac.jp}}

\date{}

\begin{document}

\maketitle

\begin{abstract}
Learning distributed representations for relation instances is a central technique in downstream NLP applications.
In order to address semantic modeling of relational patterns, this paper constructs a new dataset that provides multiple similarity ratings for every pair of relational patterns on the existing dataset~\cite{Zeichner:ACL12}.
In addition, we conduct a comparative study of different encoders including additive composition, RNN, LSTM, and GRU for composing distributed representations of relational patterns.
We also present Gated Additive Composition, which is an enhancement of additive composition with the gating mechanism.
Experiments show that the new dataset does not only enable detailed analyses of the different encoders, but also provides a gauge to predict successes of distributed representations of relational patterns in the relation classification task.
\end{abstract}

\section{Introduction}\label{introduction}
Knowledge about entities and their relations (relation instances) are crucial for a wide spectrum of NLP applications, e.g., information retrieval, question answering, and recognizing textual entailment.
Learning distributed representations for relation instances is a central technique in downstream applications as a number of recent studies demonstrated the usefulness of distributed representations for words~\cite{NIPS2013_5021,Pennington:14} and sentences~\cite{DBLP:conf/nips/SutskeverVL14,cho-EtAl:2014:EMNLP2014,Kiros:15}.

In particular, semantic modeling of relations and their textual realizations ({\it relational patterns} hereafter) is extremely important because a relation (e.g., causality) can be mentioned by various expressions (e.g., ``{\it X} cause {\it Y}'', ``{\it X} lead to {\it Y}'', ``{\it Y} is associated with {\it X}'').
To make matters worse, relational patterns are highly productive: we can produce a emphasized causality pattern ``{\it X} increase the severe risk of {\it Y}'' from ``{\it X} increase the risk of {\it Y}'' by inserting {\it severe} to the pattern.

To model the meanings of relational patterns, the previous studies built a co-occurrence matrix between relational patterns (e.g., ``{\it X} increase the risk of {\it Y}'') and entity pairs (e.g., ``{\it X}: smoking, {\it Y}: cancer'')~\cite{Lin:2001:DSI:502512.502559,Nakashole:2012:PTR:2390948.2391076}.
Based on the distributional hypothesis~\cite{harris54}, we can compute a semantic vector of a relational pattern from the co-occurrence matrix, and measure the similarity of two relational patterns as the cosine similarity of the vectors.
Nowadays, several studies adopt distributed representations computed by neural networks for semantic modeling of relational patterns~\cite{yih-he-meek:2014:P14-2,Takase:2016:MSC:2901933.2902089}.

\begin{figure}[!t]
  \centering
  \includegraphics[width=7.5cm]{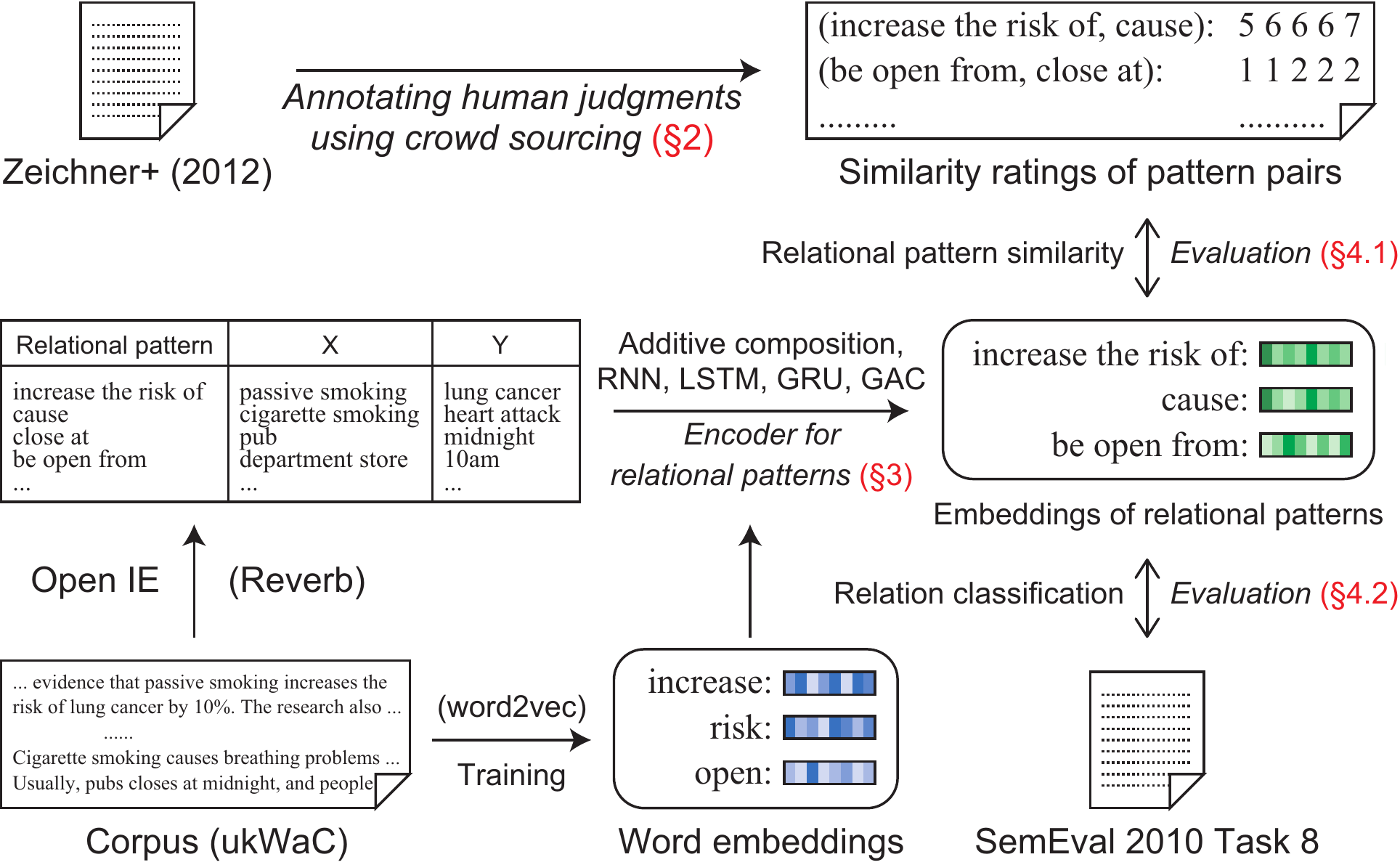}
   \caption{Overview of this study.}
   \label{researchOverview}
\end{figure}

Notwithstanding, the previous studies paid little attention to explicitly evaluate semantic modeling of relational patterns.
In this paper, we construct a new dataset that contains a pair of relational patterns with five similarity ratings judged by human annotators.
The new dataset shows a high inter-annotator agreement, following the annotation guideline of \newcite{Mitchell:Lapata:2010}.
The dataset is publicly available on the Web site\footnote{\url{http://github.com/takase/relPatSim}}.

In addition, we conduct a comparative study of different encoders for composing distributed representations of relational patterns.
During the comparative study, we present Gated Additive Composition, which is an enhancement of additive composition with the gating mechanism.
We utilize the Skip-gram objective for training the parameters of the encoders on a large unlabeled corpus.
Experiments show that the new dataset does not only enable detailed analyses of the different encoders, but also provides a gauge to predict successes of distributed representations of relational patterns in another task (relation classification).
Figure \ref{researchOverview} illustrates the overview of this study.

\section{Data Construction}
\label{sec:data}

\subsection{Target relation instances}

We build a new dataset upon the work of \newcite{Zeichner:ACL12}, which consists of relational patterns with semantic inference labels annotated.
The dataset includes 5,555 pairs\footnote{More precisely, the dataset includes 1,012 {\it meaningless} pairs in addition to 5,555 pairs. A pair of relational patterns was annotated as {\it meaningless} if the annotators were unable to understand the meaning of the patterns easily. We ignore the {\it meaningless} pairs in this study.} extracted by Reverb~\cite{fader-soderland-etzioni:2011:EMNLP}, 2,447 pairs with inference relation and 3,108 pairs (the rest) without one.

Initially, we considered using this high-quality dataset as it is for semantic modeling of relational patterns.
However, we found that inference relations exhibit quite different properties from those of semantic similarity.
Take a relational pattern pair ``{\it X} be the part of {\it Y}'' and ``{\it X} be an essential part of {\it Y}'' filled with ``{\it X} = the small intestine, {\it Y} = the digestive system'' as an instance.
The pattern ``{\it X} be the part of {\it Y}'' does not entail ``{\it X} be an essential part of {\it Y}'' because the meaning of the former does not include `essential'.
Nevertheless, both statements are similar, representing the same relation (\textsc{part-of}).
Another uncomfortable pair is ``{\it X} fall down {\it Y}'' and ``{\it X} go up {\it Y}'' filled with ``{\it X} = the dude, {\it Y} = the stairs''.
The dataset indicates that the former entails the latter probably because falling down from the stairs requires going up there, but they present the opposite meaning.
For this reason, we decided to re-annotate semantic similarity judgments on every pair of relational patterns on the dataset.

\subsection{Annotation guideline}\label{similarityData}
We use instance-based judgment in a similar manner to that of \newcite{Zeichner:ACL12} to secure a high inter-annotator agreement.
In instance-based judgment, an annotator judges a pair of relational patterns whose variable slots are filled with the same entity pair.
In other words, he or she does not make a judgment for a pair of relational patterns with variables, ``{\it X} prevent {\it Y}'' and ``{\it X} reduce the risk of {\it Y}'', but two instantiated statements ``Cephalexin prevent the bacteria'' and ``Cephalexin reduce the risk of the bacteria'' (``{\it X} = Cephalexin, {\it Y} = the bacteria'').
We use the entity pairs provided in ~\newcite{Zeichner:ACL12}.

We asked annotators to make a judgment for a pair of relation instances by choosing a rating from 1 (dissimilar) to 7 (very similar).
We provided the following instructions for judgment, which is compatible with \newcite{Mitchell:Lapata:2010}:
(1) rate 6 or 7 if the meanings of two statements are the same or mostly the same (e.g., ``Palmer team with Jack Nicklaus'' and ``Palmer join with Jack Nicklaus'');
(2) rate 1 or 2 if two statements are dissimilar or unrelated (e.g., ``the kids grow up with him'' and ``the kids forget about him'');
(3) rate 3, 4, or 5 if two statements have some relationships (e.g., ``Many of you know about the site'' and ``Many of you get more information about the site'', where the two statements differ but also reasonably resemble to some extent).

\subsection{Annotation procedure}
We use a crowdsourcing service CrowdFlower\footnote{\url{http://www.crowdflower.com/}} to collect similarity judgments from the crowds.
CrowdFlower has the mechanism to assess the reliability of annotators using Gold Standard Data (Gold, hereafter), which consists of pairs of relational patterns with similarity scores assigned.
Gold examples are regularly inserted throughout the judgment job to enable measurement of the performance of each worker\footnote{We allow $\pm 1$ differences in rating when we measure the performance of the workers.}.
Two authors of this paper annotated 100 pairs extracted randomly from 5,555 pairs, and prepared 80 Gold examples showing high agreement.
Ratings of the Gold examples were used merely for quality assessment of the workers.
In other words, we discarded the similarity ratings of the Gold examples, and used those judged by the workers.

To build a high quality dataset, we use judgments from workers whose confidence values (reliability scores) computed by CrowdFlower are greater than 75\%.
Additionally, we force every pair to have at least five judgments from the workers.
Consequently, 60 workers participated in this job.
In the final version of this dataset, each pair has five similarity ratings judged by the five most reliable workers who were involved in the pair.

\begin{figure}[!t]
  \centering
  \includegraphics[width=7.5cm]{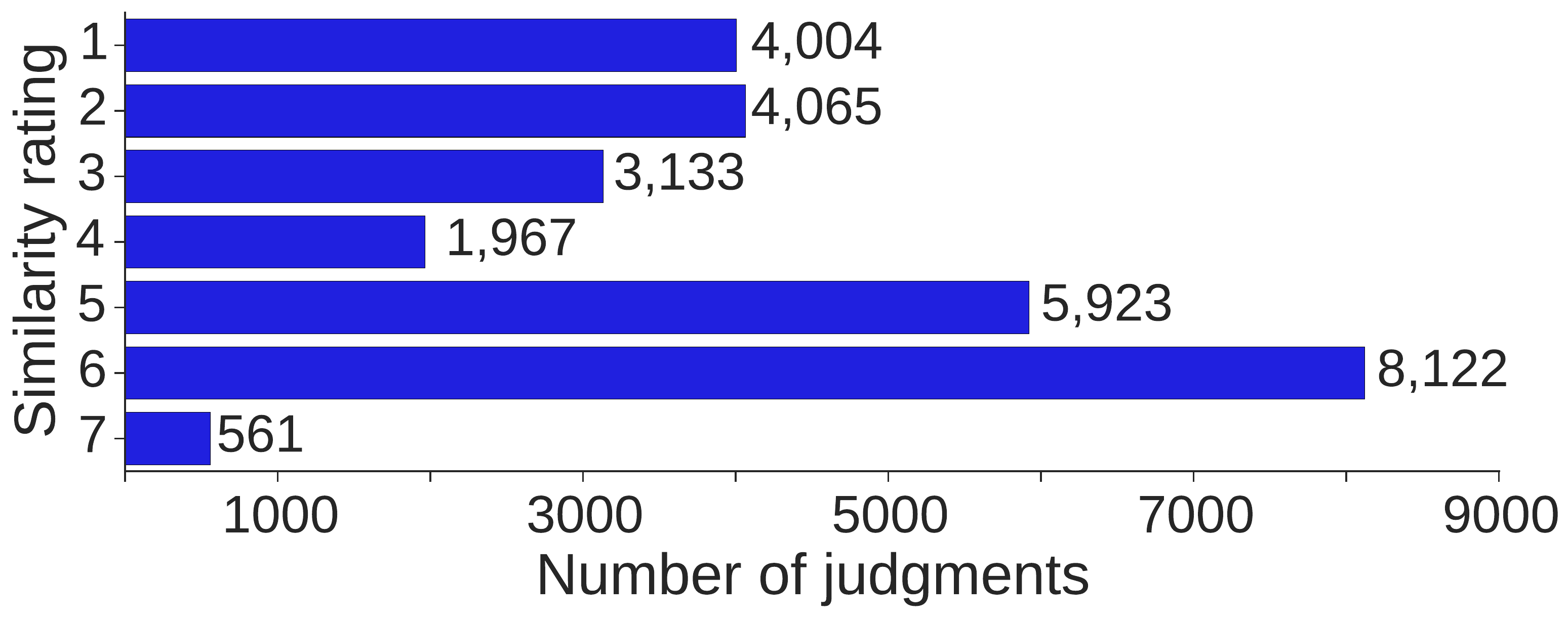}
   \caption{Number of judgments for each similarity rating. The total number of judgments is $27,775$ ($5,555 \mbox{ pairs} \times 5 \mbox{ workers}$).}
   \label{result:annotation}
\end{figure}

Figure~\ref{result:annotation} presents the number of judgments for each similarity rating.
Workers seldom rated 7 for a pair of relational patterns, probably because most pairs have at least one difference in content words.
The mean of the standard deviations of similarity ratings of all pairs is $1.16$.
Moreover, we computed Spearman's $\rho$ between similarity judgments from each worker and the mean of five judgments in the dataset.
The mean of Spearman's $\rho$ of workers involved in the dataset is $0.728$.
These statistics show a high inter-annotator agreement of the dataset.

\section{Encoder for Relational Patterns}
\label{sec:encoder}

The new dataset built in the previous section raises two new questions ---
{\it What is the reasonable method (encoder) for computing the distributed representations of relational patterns? Is this dataset useful to predict successes of distributed representations of relational patterns in real applications?}
In order to answer these questions, this section explores various methods for learning distributed representations of relational patterns.

\subsection{Baseline methods without supervision}

A na\"{i}ve approach would be to regard a relational pattern as a single unit (word) and to train word/pattern embeddings as usual.
In fact, \newcite{NIPS2013_5021} implemented this approach as a preprocessing step, mining phrasal expressions with strong collocations from a training corpus.
However, this approach might be affected by data sparseness, which lowers the quality of distributed representations.

Another simple but effective approach is {\it additive composition}~\cite{Mitchell:Lapata:2010}, where the distributed representation of a relational pattern is computed by the mean of embeddings of constituent words.
Presuming that a relational pattern consists of a sequence of $T$ words $w_1, ..., w_T$, then we let $x_t \in \mathbb{R}^d$ the embedding of the word $w_t$.
This approach computes $\frac{1}{T} \sum_{t=1}^{T} x_t$ as the embedding of the relational pattern.
\newcite{Muraoka-EtAl:2014:PACLIC} reported that the additive composition is a strong baseline among various methods.

\subsection{Recurrent Neural Network}
Recently, a number of studies model semantic compositions of phrases and sentences by using (a variant of) Recurrent Neural Network (RNN)~\cite{DBLP:conf/nips/SutskeverVL14,tang-qin-liu:2015:EMNLP}.
For a given embedding $x_t$ at position $t$, the vanilla RNN~\cite{elman1990finding} computes the hidden state $h_t \in \mathbb{R}^d$ by the following recursive equation\footnote{We do not use a bias term in this study. We set the number of dimensions of hidden states identical to that of word embeddings ($d$) so that we can adapt the objective function of the Skip-gram model for training (Section \ref{sec:parameter-estimation}).},
\begin{align}
h_t = g(W_{x}x_t + W_{h}h_{t-1}) . \label{eq:RNN}
\end{align}
Here, $W_x$ and $W_h$ are $d \times d$ matrices (parameters), $g(.)$ is the elementwise activation function ($\tanh$).
We set $h_0 = 0$ at $t = 1$.
In essence, RNN computes the hidden state $h_t$ based on the one at the previous position ($h_{t-1}$) and the word embedding $x_t$.
Applying Equation~\ref{eq:RNN} from $t=1$ to $T$, we use $h_T$ as the distributed representation of the relational pattern.

\subsection{RNN variants}
We also employ Long Short-Term Memory (LSTM)~\cite{Hochreiter:1997:LSM:1246443.1246450} and Gated Recurrent Unit (GRU)~\cite{cho-EtAl:2014:EMNLP2014} as an encoder for relational patterns.
LSTM has been applied successfully to various NLP tasks including word segmentation~\cite{ChenXinchi:15}, dependency parsing~\cite{Dyer:15}, machine translation~\cite{DBLP:conf/nips/SutskeverVL14}, and sentiment analysis~\cite{Tai:15}.
GRU is also successful in machine translation~\cite{cho-EtAl:2014:EMNLP2014} and various tasks including sentence similarity, paraphrase detection, and sentiment analysis~\cite{Kiros:15}.

LSTM and GRU are similar in that the both architectures have gates (input, forget, and output for LSTM; reset and update for GRU) to remedy the gradient vanishing or explosion problem in training RNNs.
Although some researchers reported that GRU is superior to LSTM~\cite{DBLP:journals/corr/ChungGCB14}, we have no consensus about the superiority.
Besides, we are not sure whether LSTM or GRU is really necessary for relational patterns, which ususlly consist of a few words.
Thus, we compare RNN, LSTM, and GRU empirically with the same training data and the same training procedure.
Similarly to RNN, we use the hidden state $h_T$ of LSTM\footnote{We omitted peephole connections and bias terms.} or GRU as the distributed representation of a relation pattern.

\subsection{Gated Additive Composition (GAC)}\label{explain:proposed}
\begin{figure}[!t]
  \centering
  \includegraphics[width=7.5cm]{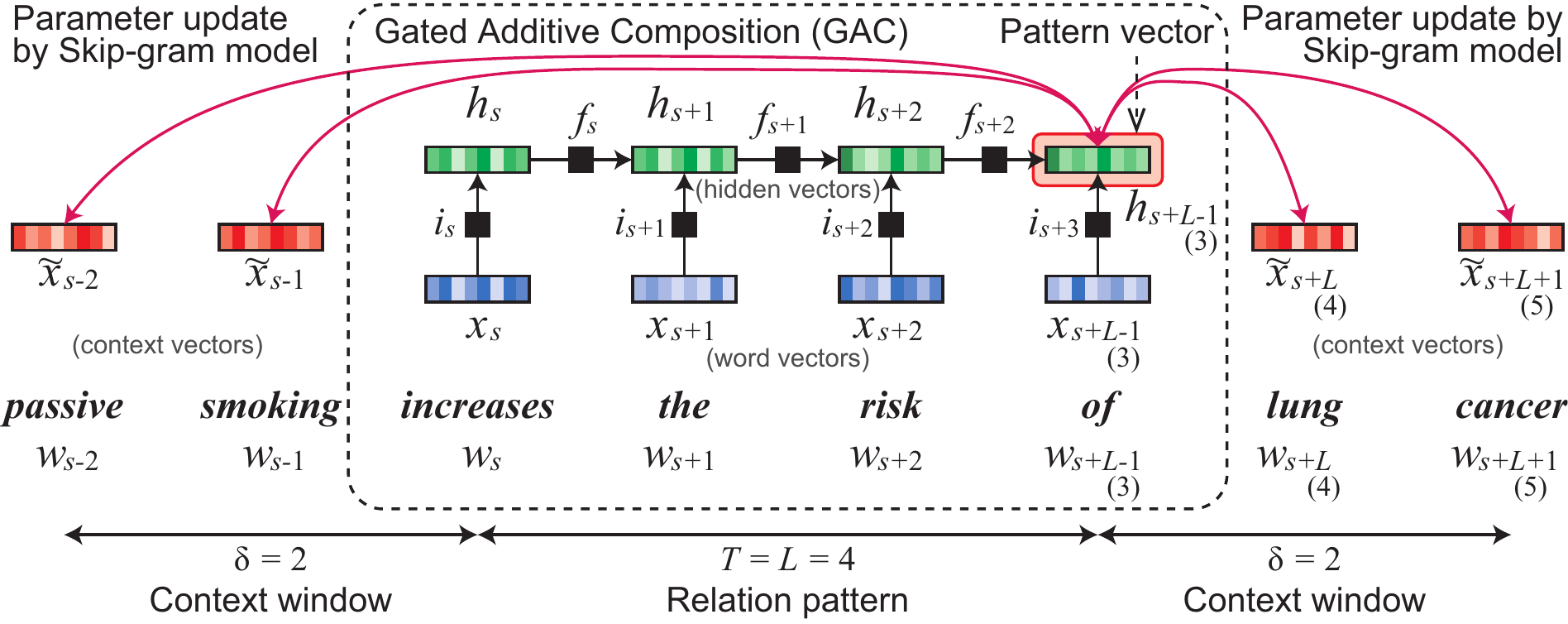}
   \caption{Overview of GAC trained with Skip-gram model. GAC computes the distributed representation of a relational pattern using the input gate and forget gate, and learns parameters by predicting surrounding words (Skip-gram model).}
   \label{illustrationOfProposedMethod}
\end{figure}

In addition to the gradient problem, LSTM or GRU may be suitable for relational patterns, having the mechanism of adaptive control of gates for input words and hidden states.
Consider the relational pattern ``{\it X} have access to {\it Y}'', whose meaning is mostly identical to that of ``{\it X} access {\it Y}''.
Because `have' in the pattern is a light verb, it may be harmful to incorporate the semantic vector of `have' into the distributed representation of the pattern.
The same may be true for the functional word `to' in the pattern.
However, the additive composition nor RNN does not have a mechanism to ignore the semantic vectors of these words.
It is interesting to explore a method somewhere between additive composition and LSTM/GRU: additive composition with the gating mechanism.

For this reason, we present an another variant of RNN in this study.
Inspired by the input and forget gates in LSTM, we compute the input gate $i_t \in \mathbb{R}^d$ and forget gate $f_t \in \mathbb{R}^d$ at position $t$.
We use them to control the amount to propagate to the hidden state $h_t$ from the current word $x_t$ and the previous state $h_{t-1}$.
\begin{align}
i_t &= \sigma(W_{ix}x_t + W_{ih}h_{t-1}) \label{eq:gac-input} \\
f_t &= \sigma(W_{fx}x_t + W_{fh}h_{t-1}) \label{eq:gac-forget} \\
h_t &= g(f_t \odot h_{t-1} + i_t \odot x_t) \label{eq:gac-hidden}
\end{align}
Here, $W_{ix}$, $W_{ih}$, $W_{fx}$, $W_{fh}$ are $d \times d$ matrices.
Equation \ref{eq:gac-hidden} is interpreted as a weighted additive composition between the vector of the current word $x_t$ and the vector of the previous hidden state $h_{t-1}$.
The elementwise weights are controlled by the input gate $i_t$ and forget gate $f_t$; we expect that input gates are closed (close to zero) and forget gates are opened (close to one) when the current word is a control verb or function word.
We name this architecture {\it gated additive composition} (GAC).

\subsection{Parameter estimation: Skip-gram model}\label{sec:parameter-estimation}
To train the parameters of the encoders (RNN, LSTM, GRU, and GAC) on an unlabeled text corpus, we adapt the Skip-gram model~\cite{NIPS2013_5021}.
Formally, we designate an occurrence of a relational pattern $p$ as a subsequence of $L$ words $w_s, ..., w_{s+L-1}$ in a corpus.
We define $\delta$ words appearing before and after pattern $p$ as the context words, and let $C_p = (s-\delta, ..., s-1, s+L, ..., s+L+\delta)$ denote the indices of the context words.
We define the log-likelihood of the relational pattern $l_p$, following the objective function of Skip-gram with negative sampling (SGNS)~\cite{Levy:14}.
\begin{gather}\label{eq:skip-gram-negative-sampling}
 l_p = \sum_{\tau \in C_p} \left(\log \sigma(h_{p}^{\top} \tilde{x}_{\tau}) + \sum_{k=1}^{K} \log \sigma(-h_{p}^{\top} \tilde{x}_{\breve{\tau}}) \right)
\end{gather}
In this formula: $K$ denotes the number of negative samples; $h_{p} \in \mathbb{R}^d$ is the vector for the relational pattern $p$ computed by each encoder such as RNN; $\tilde{x}_{\tau} \in \mathbb{R}^d$ is the context vector for the word $w_{\tau}$\footnote{The Skip-gram model has two kinds of vectors $x_t$ and $\tilde{x}_{t}$ assigned for a word $w_t$. Equation 2 of the original paper~\cite{NIPS2013_5021} denotes $x_t$ (word vector) as $v$ (input vector) and $\tilde{x}_{t}$ (context vector) as $v'$ (output vector). The {\sf word2vec} implementation does not write context (output) vectors but only word (input) vectors to a model file. Therefore, we modified the source code to save context vectors, and use them in Equation \ref{eq:skip-gram-negative-sampling}. This modification ensures the consistency of the entire model.}; $x_{\breve{\tau}'} \in \mathbb{R}^d$ is the context vector for the word that were sampled from the unigram distribution\footnote{We use the probability distribution of words raised to the 3/4 power~\cite{NIPS2013_5021}.} at every iteration of $\sum_k$.

At every occurrence of a relational pattern in the corpus, we use Stochastic Gradient Descent (SGD) and backpropagation through time (BPTT) for training the parameters (matrices) in encoders.
More specifically, we initialize the word vectors $x_t$ and context vectors $\tilde{x}_{t}$ with pre-trained values, and compute gradients for Equation \ref{eq:skip-gram-negative-sampling} to update the parameters in encoders.
In this way, each encoder is trained to compose a vector of a relational pattern so that it can predict the surrounding context words.
An advantage of this parameter estimation is that the distributed representations of words and relational patterns stay in the same vector space.
Figure \ref{illustrationOfProposedMethod} visualizes the training process for GAC.

\section{Experiments}

In Section \ref{eval:relationalPatternSim}, we investigate the performance of the distributed representations computed by different encoders on the pattern similarity task.
Section \ref{sec:relation-classification} examines the contribution of the distributed representations on SemEval 2010 Task 8, and discusses the usefulness of the new dataset to predict successes of the relation classification task.

\subsection{Relational pattern similarity}\label{eval:relationalPatternSim}

For every pair in the dataset built in Section \ref{sec:data}, we compose the vectors of the two relational patterns using an encoder described in Section \ref{sec:encoder}, and compute the cosine similarity of the two vectors.
Repeating this process for all pairs in the dataset, we measure Spearman's $\rho$ between the similarity values computed by the encoder and similarity ratings assigned by humans.

\subsubsection{Training procedure}
\label{sec:training-procedure}
We used ukWaC\footnote{\url{http://wacky.sslmit.unibo.it}} as the training corpus for the encoders.
This corpus includes the text of 2 billion words from Web pages crawled in the .uk domain.
Part-of-speech tags and lemmas are annotated by TreeTagger\footnote{\url{http://www.cis.uni-muenchen.de/~schmid/tools/TreeTagger/}}.
We used lowercased lemmas throughout the experiments.
We apply \textsf{word2vec} to this corpus to pre-train word vectors $x_t$ and context vectors $\tilde{x}_t$.
All encoders use word vectors $x_t$ to compose vectors of relational patterns; and the Skip-gram model uses context vectors $\tilde{x}_t$ to compute the objective function and gradients.
We fix the vectors $x_t$ and $\tilde{x}_t$ with pre-trained values during training.

We used Reverb~\cite{fader-soderland-etzioni:2011:EMNLP} to the ukWaC corpus to extract relational pattern candidates.
To remove unuseful relational patterns, we applied filtering rules that are compatible with those used in the publicly available extraction result\footnote{\url{http://reverb.cs.washington.edu/}}.
Additionally, we discarded relational patterns appearing in the evaluation dataset throughout the experiments to assess the performance under which an encoder composes vectors of unseen relational patterns.
This preprocessing yielded $127,677$ relational patterns.

All encoders were implemented on Chainer\footnote{\url{http://chainer.org/}}, a flexible framework of neural networks.
The hyper-parameters of the Skip-gram model are identical to those in \newcite{NIPS2013_5021}: the width of context window $\delta = 5$, the number of negative samples $K=5$, the subsampling of $10^{-5}$.
For each encoder that requires training, we tried $0.025$, $0.0025$, and $0.00025$ as an initial learning rate, and selected the best value for the encoder.
In contrast to the presentation of Section \ref{sec:encoder}, we compose a pattern vector in backward order (from the last to the first) because preliminary experiments showed a slight improvement with this treatment.

\subsubsection{Results and discussions}
\begin{figure}[t]
  \centering
  \includegraphics[width=7.5cm]{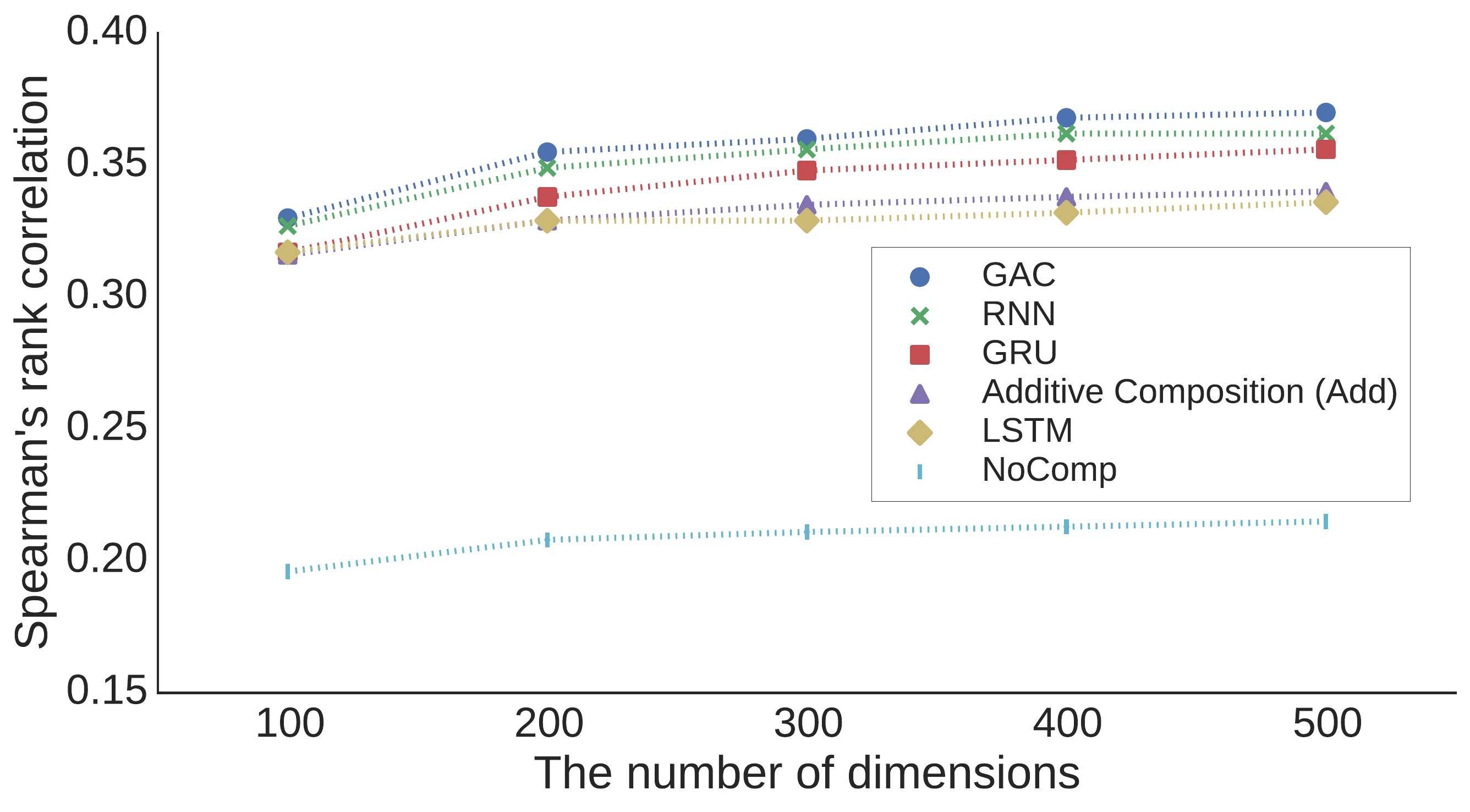}
   \caption{Performance of each method on the relational pattern similarity task with variation in the number of dimensions.}
   \label{result:patternSim}
\end{figure}

\begin{table*}[t]
  \centering
  \begin{tabular}{| r  r || r | r | r | r | r | r |} \hline
  \multicolumn{1}{|c}{Length} & \multicolumn{1}{c||}{\#} & \multicolumn{1}{|c|}{NoComp} & \multicolumn{1}{|c|}{Add} & \multicolumn{1}{|c|}{LSTM} & \multicolumn{1}{|c|}{GRU} & \multicolumn{1}{|c|}{RNN} & \multicolumn{1}{|c|}{GAC} \\ \hline
  1 & 636 & {\bf 0.324} & {\bf 0.324} & {\bf 0.324} & {\bf 0.324} & {\bf 0.324} & {\bf 0.324} \\
  2 & 1,018 & 0.215 & 0.319 & 0.257 & 0.274 & 0.285 & {\bf 0.321} \\
  3 & 2,272 & 0.234 & 0.386 & 0.344 & 0.370 & 0.387 & {\bf 0.404} \\
  4 & 1,206 & 0.208 & 0.306 & 0.314 & {\bf 0.329} & 0.319 & 0.323 \\
  $>$ 5 & 423 & 0.278 & 0.315 & 0.369 & 0.384 & {\bf 0.394} & 0.357 \\ \hline
  All & 5,555 & 0.215 & 0.340 & 0.336 & 0.356 & 0.362 & {\bf 0.370} \\ \hline
  \end{tabular}
  \caption{Spearman's rank correlations on different pattern lengths (number of dimensions $d = 500$).\label{result:spearmanForLength}}
\end{table*}

Figure~\ref{result:patternSim} shows Spearman's rank correlations of different encoders when the number of dimensions of vectors is 100--500.
The figure shows that GAC achieves the best performance on all dimensions.

Figure~\ref{result:patternSim} includes the performance of the na\"{i}ve approach, ``NoComp'', which regards a relational pattern as a single unit (word).
In this approach, we allocated a vector $h_p$ for each relational pattern $p$ in Equation \ref{eq:skip-gram-negative-sampling} instead of the vector composition, and trained the vectors of relational patterns using the Skip-gram model.
The performance was poor for two reasons:
we were unable to compute similarity values for 1,744 pairs because relational patterns in these pairs do not appear in ukWaC;
and relational patterns could not obtain sufficient statistics because of data sparseness.

Table~\ref{result:spearmanForLength} reports Spearman's rank correlations computed for each pattern length.
Here, the length of a relational-pattern pair is defined by the maximum of the lengths of two patterns in the pair.
In length of 1, all methods achieve the same correlation score because they use the same word vector $x_t$.
The table shows that additive composition (Add) performs well for shorter relational patterns (lengths of 2 and 3) but poorly for longer ones (lengths of 4 and 5+).
GAC also exhibits the similar tendency to Add, but it outperforms Add for shorter patterns (lengths of 2 and 3) probably because of the adaptive control of input and forget gates.
In contrast, RNN and its variants (RNN, GRU, and LSTM) enjoy the advantage on longer patterns (lengths of 4 and 5+).

\begin{table}[!t]
  \centering
  \begin{tabular}{| c || l | l |} \hline
        & \multicolumn{1}{|c|}{$w_t$} & \multicolumn{1}{|c|}{$w_{t+1}$ $w_{t+2}$ $...$} \\ \hline \hline
   large $i_t$ & reimburse & for \\
  (input & payable & in \\
  open)  & liable & to \\ \hline
  small $i_t$ & a & charter member of \\
  (input & a & valuable member of \\
  close) & be & an avid reader of \\ \hline
   large $f_t$ & be & eligible to participate in \\
  (forget & be & require to submit \\
  open) & be & request to submit \\ \hline
  small $f_t$ & coauthor & of \\
  (forget & capital & of \\
  close) & center & of \\ \hline
  \end{tabular}
  \caption{Prominent moments for input/forget gates.\label{result:GateHandle}}
\end{table}

To examine the roles of input and forget gates of GAC, we visualize the moments when input/forget gates are wide open or closed.
More precisely, we extract the input word and scanned words when $|i_t|_2$ or $|f_t|_2$ is small (close to zero) or large (close to one) on the relational-pattern dataset.
We restate that we compose a pattern vector in backward order (from the last to the first):
GAC scans `of', `author', and `be' in this order for composing the vector of the relational pattern `be author of'.

Table~\ref{result:GateHandle} displays the top three examples identified using the procedure.
The table shows two groups of tendencies.
Input gates open and forget gates close when scanned words are only a preposition and the current word is a content word.
In these situations, GAC tries to read the semantic vector of the content word and to ignore the semantic vector of the preposition.
In contrast, input gates close and forget gates open when the current word is `be' or `a' and scanned words form a noun phrase (e.g., ``charter member of''), a complement (e.g., ``eligible to participate in''), or a passive voice (e.g., ``require(d) to submit'').
This behavior is also reasonable because GAC emphasizes informative words more than functional words.

\subsection{Relation classification}
\label{sec:relation-classification}

\subsubsection{Experimental settings}\label{sec:relation-classification-setting}

\begin{table*}[!t]
  \centering
  \begin{tabular}{| l | l | r |} \hline
        \multicolumn{1}{|c|}{Method} & \multicolumn{1}{|c|}{Feature set} & \multicolumn{1}{|c|}{F1} \\ \hline \hline
        SVM & BoW, POS & 77.3 \\
        SVM + NoComp & embeddings, BoW, POS & 79.9 \\
        SVM + LSTM & embeddings, BoW, POS & 81.1 \\
        SVM + Add & embeddings, BoW, POS & 81.1 \\
        SVM + GRU & embeddings, BoW, POS & 81.4 \\
        SVM + RNN & embeddings, BoW, POS & 81.7 \\
        SVM + GAC & embeddings, BoW, POS & 82.0 \\
         & + dependency, WordNet, NE & 83.7 \\ \hline
        Ranking loss + GAC w/ fine-tuning & embeddings, BoW, POS & \\ 
         & + dependency, WordNet, NE & 84.2 \\ \hline \hline
        SVM~\cite{rink-harabagiu:2010:SemEval} & BoW, POS, dependency, Google n-gram, etc. & 82.2 \\ \hline
       MV-RNN~\cite{socher-EtAl:2012:EMNLP-CoNLL} & embeddings, parse trees & 79.1 \\
        & + WordNet, POS, NE & 82.4 \\ \hline
       FCM~\cite{gormley-yu-dredze:2015:EMNLP} w/o fine-tuning & embeddings, dependency & 79.4 \\
        & + WordNet & 82.0 \\
       w/ fine-tuning & embeddings, dependency & 82.2 \\
        & + NE & 83.4 \\ \hline
       RelEmb~\cite{hashimoto-EtAl:2015:CoNLL} & embeddings & 82.8 \\
        & + dependency, WordNet, NE & 83.5 \\ \hline
       CR-CNN~\cite{dossantos-xiang-zhou:2015:ACL-IJCNLP} w/ Other & embeddings, word position embeddings & 82.7 \\
       w/o Other & embeddings, word position embeddings & 84.1 \\ \hline
       depLCNN~\cite{xu-EtAl:2015:EMNLP1} & embeddings, dependency & 81.9 \\
        & + WordNet & 83.7 \\
       depLCNN + NS & embeddings, dependency & 84.0 \\
        & + WordNet & 85.6 \\ \hline
  \end{tabular}
  \caption{F1 scores on the SemEval 2010 dataset.\label{result:SemEval}}
\end{table*}

To examine the usefulness of the dataset and distributed representations for a different application, we address the task of relation classification on the SemEval 2010 Task 8 dataset~\cite{hendrickx-EtAl:2010:SemEval}.
In other words, we explore whether high-quality distributed representations of relational patterns are effective to identify a relation type of an entity pair.

The dataset consists of $10,717$ relation instances ($8,000$ training and $2,717$ test instances) with their relation types annotated.
The dataset defines 9 directed relations (e.g.,\textsc{Cause-Effect}) and 1 undirected relation \textsc{Other}.
Given a pair of entity mentions, the task is to identify a relation type in $19$ candidate labels ($2 \times 9 \mbox{ directed} + 1 \mbox{ undirected}$ relations).
For example, given the pair of entity mentions $e_1 = \mbox{`burst'}$ and $e_2 = \mbox{`pressure'}$ in the sentence ``The \textit{burst} has been caused by water hammer \textit{pressure}'', a system is expected to predict \textsc{Cause-Effect($e_2, e_1$)}.

We used Support Vector Machines (SVM) with a Radial Basis Function (RBF) kernel implemented in libsvm\footnote{\url{https://www.csie.ntu.edu.tw/~cjlin/libsvm/}}.
Basic features are: part-of-speech tags (predicted by TreeTagger), surface forms, lemmas of words appearing between an entity pair, and lemmas of the words in the entity pair.
Additionally, we incorporate distributed representations of a relational pattern, entities, and a word before and after the entity pair (number of dimensions $d=500$).
In this task, we regard words appearing between an entity pair as a relational pattern.
We compare the vector representations of relational patterns computed by the five encoders presented in Section~\ref{eval:relationalPatternSim}: additive composition, RNN, GRU, LSTM, and GAC.
Hyper-parameters related to SVM were tuned by 5-fold cross validation on the training data.

\subsubsection{Results and discussions}

Table~\ref{result:SemEval} presents the macro-averaged F1 scores on the SemEval 2010 Task 8 dataset.
The first group of the table provides basic features and enhancements with the distributed representations.
We can observe a significant improvement even from the distributed representation of NoComp (77.3 to 79.9).
Moreover, the distributed representation that exhibited the high performance on the pattern similarity task was also successful on this task; GAC, which yielded the highest performance on the pattern similarity task, also achieved the best performance (82.0) of all encoders on this task.

It is noteworthy that the improvements brought by the different encoders on this task roughly correspond to the performance on the pattern similarity task.
This fact implies two potential impacts.
First, the distributed representations of relational patterns are useful and easily transferable to other tasks such as knowledge base population.
Second, the pattern similarity dataset provides a gauge to predict successes of distributed representations in another task.

We could further improve the performance of SVM + GAC by incorporating external resources in the similar manner as the previous studies did.
Concretely, SVM + GAC achieved 83.7 F1 score by adding features for WordNet, named entities (NE), and dependency paths explained in \newcite{hashimoto-EtAl:2015:CoNLL}.
Moreover, we could obtain 84.2 F1 score, using the ranking based loss function~\cite{dossantos-xiang-zhou:2015:ACL-IJCNLP} and fine-tuning of the distributed representations initially trained by GAC.
Currently, this is the second best score among the performance values reported in the previous studies on this task (the second group of Table \ref{result:SemEval}).
If we could use the negative sampling technique proposed by~\newcite{xu-EtAl:2015:EMNLP1}, we might improve the performance further\footnote{In fact, we made substantial efforts to introduce the negative sampling technique. However, \newcite{xu-EtAl:2015:EMNLP1} omits the detail of the technique probably because of the severe page limit of short papers. For this reason, we could not reproduce their method in this study.}.

\section{Related Work}
\newcite{Mitchell:Lapata:2010} was a pioneering work in semantic modeling of short phrases.
They constructed the dataset that contains two-word phrase pairs with semantic similarity judged by human annotators.
\newcite{korkontzelos-EtAl:2013:SemEval-2013} provided a semantic similarity dataset with pairs of two words and a single word.
\newcite{TACL571} annotated a part of PPDB~\cite{ganitkevitch-vandurme-callisonburch:2013:NAACL-HLT} to evaluate semantic modeling of paraphrases.
Although the target unit of semantic modeling is different from that for these previous studies, we follow the annotation guideline and instruction of \newcite{Mitchell:Lapata:2010} to build the new dataset.

The task addressed in this paper is also related to the Semantic Textual Similarity (STS) task~\cite{agirre-EtAl:2012:STARSEM-SEMEVAL}.
STS is the task to measure the degree of semantic similarity between two sentences.
Even though a relational pattern appears as a part of a sentence, it may be difficult to transfer findings from one to another: for example, the encoders of RNN and its variants explored in this study may exhibit different characteristics, influenced by the length and complexity of input text expressions.

In addition to data construction, this paper addresses semantic modeling of relational patterns.
\newcite{Nakashole:2012:PTR:2390948.2391076} approached the similar task by constructing a taxonomy of relational patterns.
They represented a vector of a relational pattern as the distribution of entity pairs co-occurring with the relational pattern.
\newcite{grycner-EtAl:2015:EMNLP} extended \newcite{Nakashole:2012:PTR:2390948.2391076} to generalize dimensions of the vector space (entity pairs) by incorporating hyponymy relation between entities.
They also used external resources to recognize the transitivity of pattern pairs and applied transitivities to find patterns in entailment relation.
These studies did not consider semantic composition of relational patterns.
Thus, they might suffer from the data sparseness problem, as shown by NoComp in Figure \ref{result:patternSim}.

Numerous studies have been aimed at encoding distributed representations of phrases and sentences from word embeddings by using:
Recursive Neural Network~\cite{conf/icml/SocherLNM11}, Matrix Vector Recursive Neural Network~\cite{socher-EtAl:2012:EMNLP-CoNLL}, Recursive Neural Network with different weight matrices corresponding to syntactic categories~\cite{socher-EtAl:2013:ACL2013} or word types~\cite{Takase:2016:MSC:2901933.2902089}, RNN~\cite{Sutskever:11}, LSTM~\cite{DBLP:conf/nips/SutskeverVL14}, GRU~\cite{cho-EtAl:2014:EMNLP2014}, PAS-CLBLM~\cite{hashimoto-EtAl:2014:EMNLP2014}, etc.
As described in Section \ref{sec:encoder}, we applied RNN, GRU, and LSTM to compute distributed representations of relational patterns because recent papers have demonstrated their superiority in semantic composition~\cite{DBLP:conf/nips/SutskeverVL14,tang-qin-liu:2015:EMNLP}.
In this paper, we presented a comparative study of different encoders for semantic modeling of relational patterns.

To investigate usefulness of the distributed representations and the new dataset, we adopted the relation classification task (SemEval 2010 Task 8) as a real application.
On the SemEval 2010 Task 8, several studies considered semantic composition.
\newcite{gormley-yu-dredze:2015:EMNLP} proposed Feature-rich Compositional Embedding Model (FCM) that can combine binary features (e.g., positional indicators) with word embeddings via outer products.
\newcite{dossantos-xiang-zhou:2015:ACL-IJCNLP} addressed the task using Convolutional Neural Network (CNN).
\newcite{xu-EtAl:2015:EMNLP1} achieved a higher performance than \newcite{dossantos-xiang-zhou:2015:ACL-IJCNLP} by application of CNN on dependency paths.

In addition to the relation classification task, we briefly describe other applications.
To populate a knowledge base, \newcite{Riedel:13} jointly learned latent feature vectors of entities, relational patterns, and relation types in the knowledge base.
\newcite{toutanova-EtAl:2015:EMNLP} adapted CNN to capture the compositional structure of a relational pattern during the joint learning.
For open domain question answering, \newcite{yih-he-meek:2014:P14-2} proposed the method to map an interrogative sentence on an entity and a relation type contained in a knowledge base by using CNN.

Although these reports described good performance on the respective tasks, we are unsure of the generality of distributed representations trained for a specific task such as the relation classification.
In contrast, this paper demonstrated the contribution of distributed representations trained in a generic manner (with the Skip-gram objective) to the task of relation classification.

\section{Conclusion}
In this paper, we addressed the semantic modeling of relational patterns.
We introduced the new dataset in which humans rated multiple similarity scores for every pair of relational patterns on the dataset of semantic inference~\cite{Zeichner:ACL12}.
Additionally, we explored different encoders for composing distributed representations of relational patterns.
The experimental results shows that Gated Additive Composition (GAC), which is a combination of additive composition and the gating mechanism, is effective to compose distributed representations of relational patterns.
Furthermore, we demonstrated that the presented dataset is useful to predict successes of the distributed representations in the relation classification task.

We expect that several further studies will use the new dataset not only for distributed representations of relational patterns but also for other NLP tasks (e.g., paraphrasing).
Analyzing the internal mechanism of LSTM, GRU, and GAC, we plan to explore an alternative architecture of neural networks that is optimal for relational patterns.

\section*{Acknowledgments}
We thank the reviewers and Jun Suzuki for valuable comments.
This work was partially supported by Grant-in-Aid for JSPS Fellows Grant no. 26.5820, JSPS KAKENHI Grant number 15H05318, and JST, CREST.


\bibliographystyle{acl2016}

\end{document}